\documentclass[sigconf, review=false, anonymous=false, nonacm=true, screen=true]{acmart}

\AtBeginDocument{%
  \providecommand\BibTeX{{%
    \normalfont B\kern-0.5em{\scshape i\kern-0.25em b}\kern-0.8em\TeX}}}

\usepackage{hyperref}
\usepackage{cleveref}

\usepackage{array}
\usepackage{ragged2e}
\usepackage{multirow}
\newcolumntype{P}[1]{>{\centering\arraybackslash}p{#1}}
\newcolumntype{M}[1]{>{\centering\arraybackslash}m{#1}}
\newcolumntype{R}[1]{>{\RaggedLeft\arraybackslash}p{#1}}

\begin{document}

\title{Reaching the Edge of the Edge: Image Analysis in Space}

\author{Robert Bayer}
\email{roba@itu.dk}
\affiliation{%
  \institution{IT University of Copenhagen}
  \city{Copenhagen}
  \country{Denmark}
}

\author{Julian Priest}
\email{jucp@itu.dk}
\affiliation{%
  \institution{IT University of Copenhagen}
  \city{Copenhagen}
  \country{Denmark}
}

\author{P{\i}nar Tözün}
\email{pito@itu.dk}
\affiliation{%
  \institution{IT University of Copenhagen}
  \city{Copenhagen}
  \country{Denmark}
}

\renewcommand{\shortauthors}{Bayer, et al.}

\newcommand\pin[1]{\noindent{\color{red} {\bf \fbox{pin}} {\it#1}}}
\newcommand\rob[1]{\noindent{\color{red} {\bf \fbox{rob}} {\it#1}}}

\newcommand\DISCO{DISCO}

\begin{abstract}
 
Satellites have become more widely available due to the reduction in size and cost of their components. 
As a result, there has been an advent of smaller organizations having the ability to deploy satellites with a variety of data-intensive applications to run on them.
One popular application is image analysis to detect, for example, land, ice, clouds, etc. for Earth observation.
However, the resource-constrained nature of the devices deployed in satellites creates additional challenges for this resource-intensive application. 

In this paper, we present our work and lessons-learned on building an Image Processing Unit (IPU) for a satellite.
We first investigate the performance of a variety of edge devices
(comparing CPU, GPU, TPU, and VPU)
for deep-learning-based image processing on satellites.
Our goal is to identify devices that can achieve accurate results and are flexible when workload changes
while satisfying the power and latency constraints of satellites. 
Our results demonstrate that hardware accelerators such as ASICs and GPUs are essential for meeting the latency requirements. 
However, state-of-the-art edge devices with GPUs may draw too much power for deployment on a satellite.
Then, we use the findings gained from the performance analysis to guide the development of the IPU module for an upcoming satellite mission.
We detail how to integrate such a module into an existing satellite architecture and the software necessary to support various missions utilizing this module.

\end{abstract}



\keywords{earth observation, image data analysis, satellites, internet of things, edge computing, machine learning}


\maketitle

\section{Introduction}
\label{sec:intro}

In the last century, most innovations in real-world satellite applications were only available to the largest countries such as USA and Russia. 
These innovations led to significant reductions in the size of the components that make up a satellite and in the cost of the manufacturing and deployment process of a satellite.
This, in turn, introduced a new CubeSat class of small satellites.
Their format is based on a 10cm cube, with the possibility of combining multiple modules to create a larger satellite. 
This standardization \cite{cubesat} makes a batch deployment of satellites easier as the format affords tight configuration. 
The reduction in costs that came as a result of this new deployment method led to the advent of satellites owned by small or private organizations.

On the other hand, the size of the CubeSat class satellites poses new complex challenges, 
such as the power and thermal constraints as well as the physical dimensions of components. 
Therefore, space edge Internet-of-Things (IoT) applications deployed on satellites pose additional challenges compared to terrestrial edge IoT applications.
The edge devices deployed in space must perform very resource-intensive workloads with high reliability and do so in a highly resource-constrained environment.

Image processing and analysis is one class of possible satellite workloads, especially for Earth observation \cite{AksoyDZHCQDM22}. 
Satellites with this type of workload take large-scale images, which they have to store and send back to a ground station. 
The link between the satellite and a ground station is of limited bandwidth and short-lived. 
The images must be highly compressed to send all of them or filtered by quality and areas of interest.
The images already have a resolution of tens or hundreds of meters per pixel, and lossy compression would lead to even more loss of detail.
Filtering by areas of interest preserves the details of the images but is more resource-intensive.


This paper presents in detail what it takes to build a satellite with image processing capabilities
based on our experiences in a satellite mission in the 
\DISCO{} \cite{disco}
project.
More specifically, the contributions of this paper are as follows:
\begin{list}{\labelitemi}{\leftmargin=1.5em}
    \item{We characterize the requirements and constraints of an image processing unit (IPU) deployed on a CubeSat-based satellite.
    We do this by outlining multiple use case scenarios based on deep-learning-based image filtering on satellites and showing how each scenario affects the required latency, storage, and possible power draw of such a system.}
    \item{Using the identified scenarios, we investigate the suitability of several edge devices for the IPU on the satellite.
    The devices are based on a variety of hardware architectures,
    ranging from a microcontroller based on an ARM CPU to systems-on-chip (SoCs) devices offering more parallelism by leveraging the power of hardware accelerators such as graphics processing units (GPUs), tensor processing units (TPUs), and vision processing units (VPUs).
    Our results demonstrate not only the benefit of using highly specialized hardware 
    but also the need to strike a balance across flexibility, energy efficiency, and image processing latency for our space edge IoT use case.}
    \item{Based on our analysis of the edge devices, we identify the CoralAI Mini with the TPU as the most suitable one for the target application and
    illustrate what is necessary to integrate it within an existing CubeSat to create an image processing module on the satellite.
    In parallel, we highlight the reliability pitfalls of software running on remote deployments, such as satellites, and how to overcome them.}
\end{list}

The rest of the paper is organized as follows.
\Cref{sec:background} surveys related work and introduces the \DISCO~project.  
\Cref{sec:requirements} lists the requirements specific to deploying an IPU on a CubeSat based on three use case scenarios,
while \Cref{sec:ipu} experimentally evaluates a variety of edge devices under these use cases and requirements. 
Based on the results of our evaluation,
\Cref{sec:system} describes the IPU we built and integrated into the CubeSat-based satellite for the \DISCO~project
while discussing design alternatives and what to pay attention to in future designs. 
Finally, we conclude in \Cref{sec:conc}.

\section{Background}
\label{sec:background}

Before identifying the requirements for the image processing unit on the satellite, 
this section first surveys related work on data-intensive applications at the edge with a specific focus on satellites
and introduces the \DISCO~project.


\subsection{Related Work}
\label{sec:background:related}

Earth observation or other types of satellite imagery are amongst the most common data-intensive workloads on satellites. 
The satellite is a source of image data.
The challenge of low bandwidth on small satellites emphasizes the need to post-process that image data on-satellite to minimize what needs to be sent to the ground. 
The combination of the size of the images and the lack of computing power in the past prohibited a heavy use of on-satellite image processing or machine learning.
With the evolution of modern hardware in recent years, this has changed, though.
Now there is an influx of specialized hardware for machine learning at the edge:
NVIDIA Jetsons \cite{BayerTT22}, Google CoralAI \cite{coral-tpu}, Intel Neural Stick \cite{compute-stick}, etc.
Furthermore, the computing and memory resources available on edge devices has been increasing. 
As a result, multiple works have investigated solutions for accelerating on-satellite data post-processing, such as image classification or segmentation, on the limited power budget this environment requires.

While microcontrollers or more complex CPUs have extensive flight history (have already been deployed in satellites before \cite{aau-sat, tokyo-sat}) and low power footprint,
they were not built with running deep learning workloads in mind.
In contrast, system-on-chip (SoC)-based devices using small GPUs, such as the NVIDIA Jetsons, have been explored more in the past decade in satellites \cite{dnncubesat, accelerating_space, broncosat}
with one of the leading workloads being machine learning,
based on the success of GPUs in terrestrial use cases of machine learning. 
While CPUs and GPUs offer more flexibility than application-specific integrated circuits (ASICs),
ASICs for machine learning have also seen deployments on satellites recently
since they are designed to perform fast neural network inference, promising low latency with high efficiency.
Examples of such deployments have been with the multiple versions of Intel’s Movidius Myriad vision processing unit (VPU) \cite{esa_remote_sensing, cloudscout},
and the newer CoralAI Edge TPUs \cite{Goodwill21}.
Nevertheless, to the best of our knowledge,
there has not been a thorough benchmarking of this variety of devices in the context of image processing on satellites like the one we perform in \Cref{sec:ipu}.

Specialized devices with field-programmable gate arrays (FPGAs) have also been explored for the acceleration of deep learning inference on satellites \cite{fpgabenchmarkpositionpaper}.
While we also investigated FPGAs at the beginning of our project,
we later deemed them unsuitable for the requirements of \DISCO~(see \Cref{sec:system:discussion})
and, therefore, omitted them from the experimental analysis in \Cref{sec:ipu}.


In addition to the focus on satellite applications,
the data management community has put effort into data management and processing for terrestrial IoT and edge computing.
This includes works benchmarking machine learning performance for the edge \cite{BayerTT22, benchmarking-efficient-accelerators, benchmarking-tinyml, mlperftiny},
optimization of deep learning inference on resource-constrained devices \cite{logicnets, dl-on-mcu},
data management and processing on embedded and resource-constrained environments \cite{duckdb-cidr, picoDB, mobileDBsurvey},
data handling and processing for streaming data \cite{nebulastream, sparqliot, ChatziliadisZZM21, frontier},
and managing satellite imagery on the ground \cite{AksoyDZHCQDM22}.
We complement these works with a very specific application focus, which is data-intensive applications deployed on satellites.
Furthermore, we detail the full system description for a real-world satellite deployment in \Cref{sec:system}.

\subsection{\DISCO}
\label{sec:background:disco}


The \DISCO~program offers university students the opportunity to design and operate a small satellite and to gain space flight experience. 
It is a collaboration between four universities and aims at launching three student CubeSats into Low Earth Orbit.
The first launch, including the units we designed and describe in \Cref{sec:system}, has already happened in April 2023.

The first satellite (\DISCO1) is a 1U CubeSat, which has two main purposes: (1) to serve as a proof of concept for the image processing unit (IPU) in its payload, and (2) to provide a platform for running student projects in areas such as tiny machine learning \cite{tflitemicro} and attitude determination and control.
This satellite is the one that was part of the first launch mentioned above. 

The second, larger, 3U satellite (\DISCO2) has been designed by students, including the ones in our team, to carry an \textit{Earth imaging payload}.
This payload is an infrastructure to capture images of the Earth from a sun-synchronous polar orbit\footnote{Orbit in which the satellite passes over any given point of the planet's surface at the same local solar time.}. 
The initial focus for the project is the long-term marine systems and cryosphere\footnote{The surface on Earth where the water is in solid form such as snow and ice.} monitoring projects in the Arctic region. 
Overall, the satellite will supplement ground-based observations with remote sensing data, providing both good polar coverage and on-demand availability for field research projects.
The payload of this satellite will include up to three high-resolution cameras, as well as a dedicated IPU capable of performing image processing on images coming from these cameras, representing a highly data-intensive workload.

\subsubsection{Architecture Overview}
\label{sec:background:disco:arch}

\DISCO1 and \DISCO2 follow a very similar structure. 
Both are built of modules that can communicate with each other thanks to the inclusion of a microcontroller in each of these modules. 
The communication is facilitated by a CubeSat Space Protocol (CSP) \cite{csp}, a light protocol stack designed around the TCP/IP model, featuring transport and routing protocols and MAC-layer interfaces. 
This design, i.e., the inclusion of the microcontroller at each module, reduces a single point of failure, increases redundancy, and minimizes dependencies among the modules.

CSP provides means to send packets with arbitrary payloads between the nodes
(e.g., between modules on satellite, between ground and satellite).
It is heavily focused on a service-based architecture, which can be defined for each device specifically.
CSP also provides other general-purpose services, such as ping, reboot, or shutdown. 
In addition to services, each node also exposes parameters through its own parameter table, which enables asynchronous processing of requests between these nodes. 
The parameters can be retrieved and changed by any of the devices on the network. 
These parameters can include telemetry data, such as temperature, power consumption, or uptime
in addition to flags for triggering actions, e.g., capturing an image, results of these actions, etc.

The CSP stack is extensible to any networking interface, the CAN bus \cite{can} and KISS \cite{kiss} (RS422 \cite{rs422}) being the most commonly used ones and included out-of-the-box.
Their commonality is thanks to their suitability for deployments in harsh environments as they decouple the 
modules to protect them from high voltage transient events, such as electrostatic discharge. 
In addition to this basic functionality, CSP provides means for HMAC-SHA1 \cite{hmac} authentication and CRC32 error checking \cite{crc32}, used for communication between the satellite and the ground station. 
The satellite segment of the network is primarily run on FreeRTOS \cite{freertos}, while the ground segment runs on general-purpose Linux,
which is also used within the satellite segment on rare occasions.

\begin{figure}[h!]
\centering
\includegraphics[width=0.9\linewidth]{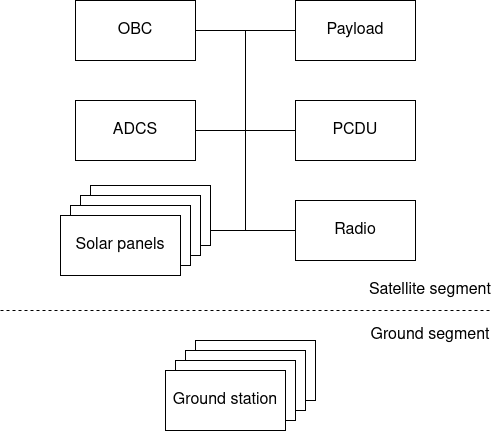}
\caption{Block diagram of the satellite and ground segments of the CSP network of the \DISCO~deployments.}
\label{fig:network_architecture}
\end{figure}

\subsubsection{Modules}
\label{sec:background:disco:modules}

\vspace{-1.5em}
Figure \ref{fig:network_architecture} shows the block diagram of the satellite and ground segments. 
The satellite segment shows the modules the \DISCO~satellites are composed of. 
While the high-level architecture of both satellites is the same, the following description of the modules will mainly focus on the deployment of the \DISCO1 satellite to simplify the descriptions.

\paragraph{On-board Computer (OBC) \cite{obc}} 
This is the most general-purpose module of the satellite containing a redundant ARM Cortex-M7, the most capable microcontroller outside the \textit{payload}. 
This module performs mission-critical tasks such as attitude determination and control and runs experiments using a variety of sensors (magnetometers, temperature sensors, light sensors, etc.) on board.

\paragraph{Radio \cite{ttc}} 
A full-duplex radio facilitates communication between the satellite and ground segments of the network. 
The radio on board of \DISCO1 is only $9600$ baud and can therefore achieve a maximum speed of $9600$ bit/s, while the S-band radio on \DISCO2 will be able to achieve a downlink rate of up to 10Mbit/s, with the same uplink rate.
This module is also responsible for possible authentication and error checking of messages.

\paragraph{Attitude Determination and Control System (ADCS)} 
The most common instruments for changing orientation or performing a possible detumbling\footnote{Satellites can start spinning uncontrollably after deployment. Detumbling is the process of stabilizing their orientation.} of a satellite are reaction/momentum wheels and magnetorquers.
The reaction wheels use motors with attached fly wheels.
The change in the angular velocity of the fly wheel causes the satellite to counter-rotate through the conservation of angular momentum.
The magnetorquers use magnetic coils to push against the Earth's magnetic field and therefore orient themselves with respect to Earth.
The position of the satellite with respect to the Earth is determined using magnetometers or using the sun sensors with regards to the Sun.
\DISCO1 uses two magnetorquers to possibly detumble itself, which are part of the solar panels.

\paragraph{Solar panels}
The whole surface of the satellite is covered by solar panels, responsible for the generation of power for the satellite. 
Additionally, two of the panels include magnetorquers, while two other panels contain retracted antennas, which deploy once in orbit.

\paragraph{Power Conditioning and Distribution Unit (PCDU)}
The electricity harvested using solar panels is stored in batteries and distributed to other modules using PCDU.
This module is also responsible for monitoring the use of power and cut off modules in case of unexpected spikes or to preserve the charge of the battery.

\paragraph{Payload}
Payload is the module that performs image processing and analysis on the satellite.
\DISCO1's payload contains the IPU as a proof-of-concept for the payload of the \DISCO2 satellite, where it will support image processing and subsequent machine learning on images coming from at least one camera during Earth observation. 
While \DISCO1 does not contain any outward-facing cameras, we will use the IPU on board of \DISCO1 to run tests to prove the suitability of the IPU chosen in this work for use on \DISCO2. 
Next, we outline the requirements for the IPU. 

\section{Requirements}
\label{sec:requirements}

The requirements and constraints of the system are based on the \DISCO2 Arctic imaging mission use case. 
The power and mass constraints of the IPU are based on the budget and engineering design of \DISCO2.
These are typical values for a modular 3U Earth-imaging satellite. 
One third of the satellite (1U) is dedicated for the \textit{payload} to house the IPU as well as the module enclosure, cameras, and corresponding optics. 
The mass of the payload cannot exceed $1.3$kg.
Table \ref{tab:constraints} gives the full list of the IPU's physical constraints.


\begin{table}[h]
\centering
\begin{tabular}{lr}
\toprule
Nominal Power & 2.00 W \\ 
Peak Power  & 5.00 W \\ 
Power & 1.48 W \\ 
Design Margin & 5\% \\ 
Nominal Cycle & 20\% \\ 
Peak Cycle & 20\% \\ 
Mass budget  & 0.15 Kg \\ 
Dimensions & 10x70x80 mm \\
\bottomrule
\end{tabular}
\caption{List of constraints of the Image Processing Unit (IPU) on board of the \DISCO2 satellite.}
\label{tab:constraints}
\vspace{-3em}
\end{table}

In addition to the physical constraints, the IPU should meet soft requirements.
Such requirements can be short time-to-boot, robustness against crashes and failures, and preferably flight heritage (history of successful deployments of a component).
Furthermore, there are use-case-related requirements, such as latency, which we base on a setup with a single streaming camera, which can be extrapolated to multi-camera setups.
The IPU must also have low heat dissipation as it has to rely purely on radiation for cooling, as convection is almost non-existent in space.

The planned orbit of \DISCO2 is a sun-synchronous polar orbit at 550 km altitude. 
The camera sensor is based on an Alvium 1800 C-2040 with an image resolution of 4512 x 4512 pixels, and the lens focal length is the maximum that could be considered for the mission. 
Assuming a 50\% overlap between images, i.e., capturing the same land area, this provides a minimum time of 4.42s between consecutive images ($T_i$), which is derived as follows:

\begin{equation}
\label{eq:latency}
T_i = \frac{GSD * H_i * \cap_i}{v} = \frac{14.8495 m/px * 4,512 px * 0.5} {7,585.16 m/s} = 4.42s
\end{equation}

\noindent , where $GSD$ (ground sample distance) is the spatial resolution of an image,
$H_i$ is the height of the image in pixels,
$\cap_i$ is the overlap between images, and
$v$ is the orbital velocity of the satellite.

Given this latency, we next describe three image processing scenarios with varying levels of difficulty based on the Arctic mission's goals.
For multi-camera setups, the substitution of other cameras' $GSD$ and $H_i$ would determine the latencies for each of them.
The overall latency requirement for the IPU would then depend on whether the cameras are used simultaneously or not.

\paragraph{Scenario 1: Real-time imaging.}
Images are taken with a period of $4.42$s, as derived in Equation \ref{eq:latency}.
The inference has to be performed and completed before the next image is captured.
This would allow for the results of an inference to be used in decision-making for the next image capture.

\paragraph{Scenario 2: Arctic region imaging.}
In this scenario, images are taken only while the satellite is over the polar regions relaxing the latency requirements.
Images are buffered and inference is performed in the remainder of the orbit,
with the IPU available for inference before the subsequent orbit.
The capture-inference cycle completes in $5,739$s, which, given the camera's spatial resolution, amounts to the capture of $80$ images and $71.74$s per image.

\paragraph{Scenario 3: Greenland imaging.}
The images are taken only while over Greenland, further relaxing the latency requirements.
As the orbit is sun-synchronous with a period of one day, this results in subsequent bursts of images being taken but only when the swathe passes over Greenland.
There are four consecutive passes over Greenland due to its large width. 
Each pass takes 80 images ($N_i$), which can be derived as follows:
\begin{equation}
    N_i = \frac{H_G}{GSD * H_i * \cap_i} = \frac{2,670,000m}{14.8495 m/px * 4,512px * 0.5} = 79.7
\end{equation}
, where $H_G$ is the distance between the northernmost and the southernmost points of Greenland.
As a result, $80 x 4 = 320$ images have to be buffered for performing inference on them.
This yields the maximum per-image inference latency of 270s
(number of seconds in a day / number of images).

While this dramatically decreases the compute requirements, it increases the requirements for the storage capacity and I/O, as the images have to be first buffered to storage and then read for inference and possibly deleted if the image is deemed unimportant for the mission.

\section{Choice of the IPU}\label{sec:ipu}

 With the requirements for the IPU identified in \Cref{sec:requirements},
this section investigates the suitability of a list of edge devices as a choice for the IPU.
The list of devices represent different processing types and capabilities at edge as described in \Cref{sec:ipu:setup}.
As a result, there is no one size fits all when it comes to the software setup required for the target use case on these devices.
To ensure that we utilize them in the best way possible for the target workload,
we first analyze the performance trade-offs
and settle on the most optimized setup for each of them
in \Cref{sec:ipu:optimizations},
before we compare the performance of all devices with respect to each other in \Cref{sec:ipu:results}.
Finally, \Cref{sec:ipu:summary} gives a summary of our investigation.


\subsection{Methodology and Setup}\label{sec:ipu:setup}

\subsubsection{Devices under Test}\label{sec:ipu:setup:devices}

We conduct experiments on six devices representing different hardware architectures, from general-purpose to ASICs. 
The goal is to show the degree of specialization needed to perform the role of the IPU on board of a satellite, as well as the efficiency these devices can achieve while doing so. 
In addition to the variety of hardware architectures to represent,
the devices are chosen based on their physical dimensions and mass
while taking into account the need for high performance and the highly limited power budget.

\begin{table}[h]
\centering
\begin{tabular}{@{}lr@{}}
\toprule
Processor & ARM Cortex-M7 @ 300 MHz \\
SRAM      & 384 KB SRAM                \\
FRAM      & 32 KB                      \\
Flash     & 2 MB                       \\
Storage   & 64 GB eMMC               \\
Dimensions& 94 x 94 x 13 mm        \\
Mass    & 120 g                \\ 
\bottomrule
\end{tabular}
\caption{Specifications of the OBC-P3 system containing the ARM Cortex-M7 MCU. The dimensions and weight includes the enclosure with aluminium shielding.}
\vspace{-2em}
\label{tab:arm}
\end{table}

\paragraph{ARM Cortex-M7 Microcontroller \cite{obc}} 
This device is the most general-purpose one on our list since it only has ARM CPUs.
It delivers the lowest performance due to its low specialization. 
It, however, has a low power consumption, which is an important factor in the choice of the IPU. 
As an additional advantage, this chip has an extensive flight history and would therefore be a safe choice for deployment in space applications.

To perform machine learning inference using this microcontroller (MCU), we leverage TensorFlow Lite for Microcontrollers \cite{tflitemicro},
which is a framework designed to perform inference with a low memory overhead without a need for an operating system.

When experimenting with this device option,
we emulate the performance and power characteristics of a flight computer already available for potential development in the \DISCO~project (OBC-P3).
This is done using the Cortex-M7 core of the \textit{STM32H745} chip and scaling its clock frequency down to 300MHz.
The full set of parameters of the OBC-P3 system can be found in Table \ref{tab:arm}.

\begin{table}[h]
\centering
\begin{tabular}{@{}lr@{}}
\toprule
CPU & Quad-core ARM A57 @ 1.43 GHz \\
GPU      & 128-core Maxwell \\
GFLOPS     & 472 \\
RAM      & 4 GB 64-bit LPDDR4 25.6 GB/s \\
Storage     & 64 GB SD card \\
Dimensions  &   100 x 80 x 29 mm    \\
Mass  &   141 g               \\
\bottomrule
\end{tabular}
\caption{Specifications of the NVIDIA Jetson Nano. The dimensions and weight are based on the developer kit version.} 
\vspace{-2em}
\label{tab:nano}
\end{table}

\paragraph{NVIDIA Jetson Nano \cite{nano}} 
This device is designed for embedded applications built around a power-efficient SoC composed of an ARM CPU and an NVIDIA GPU. 
GPUs are the most common choice for the acceleration of deep learning inference thanks to the high degree of parallelism they offer. 
This device, however, is not only specialized for deep learning inference but also can be leveraged to accelerate other parallel tasks, such as the ones found in image pre-processing pipelines (cropping, resizing, etc.). 
This flexibility, while allowing a more general-purpose use, can have increased power consumption as one of its side effects.

NVIDIA Jetson Nano can operate at a wattage between 5W - 10W but was configured to operate at 5W by disabling two of the four cores of the CPU to fit within the power budget found in Table \ref{tab:constraints}. 
Table \ref{tab:nano} shows the full specifications of this device.

This device runs Ubuntu 18.04.
Since the device is flexible, the deployment of inference on this device can leverage frameworks such as TensorFlow \cite{tensorflow2015-whitepaper}, PyTorch \cite{torch}, or TensorRT \cite{trt}.

\begin{table}[h]
\centering
\begin{tabular}{@{}lr@{}}
\toprule
CPU & Quad Core Broadcom BCM2837 @ 1.2GHz \\
RAM      & 1 GB \\
Storage     & 32 GB SD card \\
Dimensions  &   100 x 80 x 29 mm    \\
Mass  &   42 g               \\
\bottomrule
\end{tabular}
\caption{Specifications of the Raspberry Pi model 3.}
\vspace{-2em}
\label{tab:rpi}
\end{table}
\begin{table}[h]
\centering
\begin{tabular}{@{}lr@{}}
\toprule
Interface & USB 3.1, USB 2.0 \\
Dimensions & 72.5 x 27 x 14 mm \\
Mass & 53.4 g \\
\bottomrule
\end{tabular}
\caption{Specifications of the Intel Neural Compute Stick 2.}
\vspace{-2em}
\label{tab:ncs2}
\end{table}

\paragraph{Intel Neural Compute Stick 2 (NCS2) \cite{compute-stick}}
This device is an ASIC based on the Intel Movidius Myriad X Vision Processing Unit (VPU),
which is designed to accelerate convolutional neural network inference while delivering low power consumption. 
This accelerator needs to be coupled with a host device, which in the case of this study is Raspberry Pi 3 \cite{rpi}. 
The specifications of the host device and Intel NCS2 are shown in Tables \ref{tab:rpi} and \ref{tab:ncs2}, respectively.

The deployment of neural network inference on NCS2 is facilitated by the Intel OpenVino framework \cite{openvino}, which can convert the neural networks trained using a majority of popular deep learning frameworks into its intermediate representation. 
OpenVino further optimizes these models and compiles them for running on the NCS2. 
These optimizations include layer fusion and quantization to 16-bit floating point values, the only data type supported by NCS2.

\begin{table}[h]
\centering
\begin{tabular}{@{}lR{9.5em}R{9.5em}@{}}
\toprule
&   CoralAI Dev Board Micro  &   CoralAI Dev Board Mini\\
\midrule
CPU & ARM Cortex-M7 @ 800 MHz, & Quad-core ARM Cortex-A35 @ 1.5 GHz  \\
    & ARM Cortex-M4 @ 400 MHz  &   \\
RAM      & 64MB & 2 GB  \\
Storage      & 128MB NAND & 8 GB eMMC   \\
Dimensions  & 65.0 x 30.0 x 6.8 mm &   64 x 48 x 14.6 mm \\
Mass  & 10.4 g &  25.5 g\\
\bottomrule
\end{tabular}
\begin{tabular}{@{}m{6.5em}R{3em}R{4.3em}R{5.5em}R{2.5em}@{}}
    & TOPS & Interface &   Dimensions  &   Mass \\
    \midrule
    CoralAI USB          & 4    &   USB2    &   65 x 30 mm   &   4.3 g \\
    accelerator & & & & \\
    \bottomrule
\end{tabular}
\caption{Specifications of the CoralAI Dev Board Micro and CoralAI Dev Board Mini, as well as the CoralAI USB accelerator. The CoralAI Dev Board Mini's and CoralAI Dev Board Micro's physical dimensions and weight include the on-board TPU, and TOPS is tera operations per second.}
\vspace{-2em}
\label{tab:tpu}
\end{table}

\paragraph{CoralAI TPU \cite{coral-tpu}}
This device is an AI accelerator developed by Google to accelerate machine learning inference at the edge. 
Similarly to the NCS2, this accelerator is not a standalone device and needs to be coupled with a host device. 

The TPU compiler \cite{tpu_compiler} accepts neural networks in the TensorFlow Lite \cite{tflite} format, quantized to 8-bit integers, the device's only supported data type. 

This device was evaluated in three form factors:
\begin{enumerate}
    \item Coral AI Dev Board Micro \cite{coral-micro}, an embedded device coupling the CoralAI TPU with two low-power ARM cores running FreeRTOS \cite{freertos}.
    \item Coral AI Dev Board Mini \cite{coral-mini}, which combines the CoralAI TPU with a more capable quad-core CPU at the cost of higher power consumption.
    \item Coral AI USB Accelerator \cite{coral-usb}, hosted on the Raspberry Pi Model 3 (\Cref{tab:rpi}), which will provide a fair comparison between this device and the NCS2.
\end{enumerate}
\Cref{tab:tpu} details the relevant specifications for all three form factors. 

\subsubsection{Metrics}\label{sec:ipu:setup:metrics}

The metrics we consider during the evaluation of the performance of these devices are based on the requirements outlined in \Cref{sec:requirements}.

\paragraph{Latency}
Reported in seconds, this metric measures the time-to-inference of an image of size 4512 x 4512 pixels,
since this is the image resolution of our camera (\Cref{sec:requirements}). 
The latency considers the case where the image is already in the device's memory and disregards the latency of image capture, network overhead, and setup time, including the loading of the models, which will get amortized over time in a real-world setting.

\paragraph{Nominal power draw}
This is the average power draw of the device multiplied by the device's \textit{duty cycle}, which is the ratio between the achieved and required latency for each scenario outlined in \Cref{sec:requirements}. 
This metric is reported in \textit{mW}. 

\paragraph{Peak power draw}
This is the maximum power draw, measured in \textit{mW}, a device reaches during the inference.

\paragraph{Power consumption}
Reported in \textit{mWh},
Equation \ref{eq:power consumption} shows how we derive power consumption (E), where $\bar{P}$ is the average power draw in mW and $t$ is the latency in seconds of inference of a single full-size image.
This metric will guide the choice of a more energy-efficient device if multiple devices fulfill the requirements for a particular use case scenario.
\begin{equation}\label{eq:power consumption}
    E=\bar{P}(t/3,600)
\end{equation}

The power draw of the devices was measured using Otii Arc Pro \cite{otii}, a precision power supply and analyzer.

\paragraph{Accuracy}
This metric is tracked to quantify the effects the different computational optimizations, such as quantization, have on the predictive performance of the deployed models.

\subsubsection{Workload}\label{sec:ipu:setup:workload}

\paragraph{Model and Dataset}
To simulate the imaging scenarios outlined in Section \ref{sec:requirements}, we use a 5-class image classification problem aligned with one of the initial tasks the satellite will run - land cover classification.
For this, we fine-tuned a MobileNetV1 \cite{mobilenet} neural network on the Flowers dataset \cite{tfflowers}. 
The choice of the model was guided by its small size and high flexibility in scaling down the size of the model using the \textit{depth multiplier},
a parameter for controlling the number of convolutional layers of the network.
The scaling down is necessary to fit this model on the smallest of the devices under test, the ARM Cortex-M7 MCU, which only has a little over 2MB of memory. 
While the MobileNetV1 model has two successors, versions 2 and 3, neither of them can scale down as far as the first version.

The model was pre-trained on the ImageNet dataset \cite{imagenet} and fine-tuned on the Flowers dataset, which matched our use case in terms of the number of classes and the expected input size of the MobileNet models (224 x 224 pixels).
While measuring the reported \textit{accuracy}, we use the Flowers dataset to quantify the possible loss in predictive performance due to optimizations such as quantization.

\paragraph{Inference scenario}
The contents of the Flowers dataset are not representative of the images the IPU will process during its deployment.
However, the shape of the images in the dataset is the most important factor for our analysis when reporting \textit{latency} and \textit{power}-related metrics, since the shape, not the content, of an image affects the computational requirements for the inference.

To emulate the computational requirements of the expected inference scenario,
we randomly generate a 4512 x 4512 image ($\sim$60MB) following the size of the images produced by our camera (\Cref{sec:requirements}).
This image is essentially a matrix with dimensions 4512x4512x3.
The image is then partitioned into tiles of 224 x 224 creating 400 patches from the original image, in order to fit into the memory space of all the devices under test and the model input size.\footnote{The image is not evenly divisible. Therefore, we disregard the last 32 rows and columns of the image matrix.}
We perform the inference on these patches separately. 
The reported latency is the latency to perform the inference on all 400 patches. 
This method furthermore serves as a coarse-grain image segmentation in practice, since it can allow us to send only the patches interesting to the mission at hand rather than the whole 4512 x 4512 image, leading to further optimization in bandwidth utilization between the satellite and the ground.

Finally, the inference does not include any image pre-processing, except for the necessary image rescaling to the range of $[0,1)$ in the case of the NVIDIA Jetson Nano and the NCS2, which operate on floating point values rather than integers.

\paragraph{Runs}
The performance of the devices is measured for ten inferences of the full-sized images.
The results reported are the average values from these ten runs. 
All devices are tested on all three sizes of the network (depth multiplier of 0.25, 0.5, and 1.0), with the exception of the ARM Cortex-M7 MCU, which could only fit the smallest of the models in memory as also mentioned above.

\paragraph{Code}
The code\footnote{The codebase can be found at \url{https://github.com/rbcarlos/Reaching-the-Edge-of-the-Edge-Image-Analysis-in-Space}} outside the used frameworks is written in C++ for higher performance and efficiency, with the exception of the code for Jetson Nano, which is written in Python. The use of Python in the case of Jetson Nano does not affect the inference latency significantly, as the majority of the time is spent on the GPU. 

\subsection{Optimizations}\label{sec:ipu:optimizations}

To take full advantage of the variety of the hardware architectures at hand, we apply multiple optimizations for each setup prior to comparing them against each other.

\subsubsection{ARM Cortex-M7 Microcontroller}\label{sec:ipu:optimizations:mcu}

While the MCU has support for floating point operations, we quantized the network to 8-bit precision to increase the performance and, more importantly, decrease the memory requirements of running the models because the memory size is the most limiting factor of this setup.

TensorFlow Lite for microcontrollers allows for the use of specialized neural network kernels. 
We leverage the CMSIS-NN kernels \cite{cmsis_nn} that are specifically designed and optimized to run on the ARM Cortex-M class of devices.
The X-CUBE-AI framework \cite{xcubeai} is used to simplify the deployment of the model with the custom kernels. 

\begin{figure}[ht]
\centering
\includegraphics[width=0.95\linewidth]{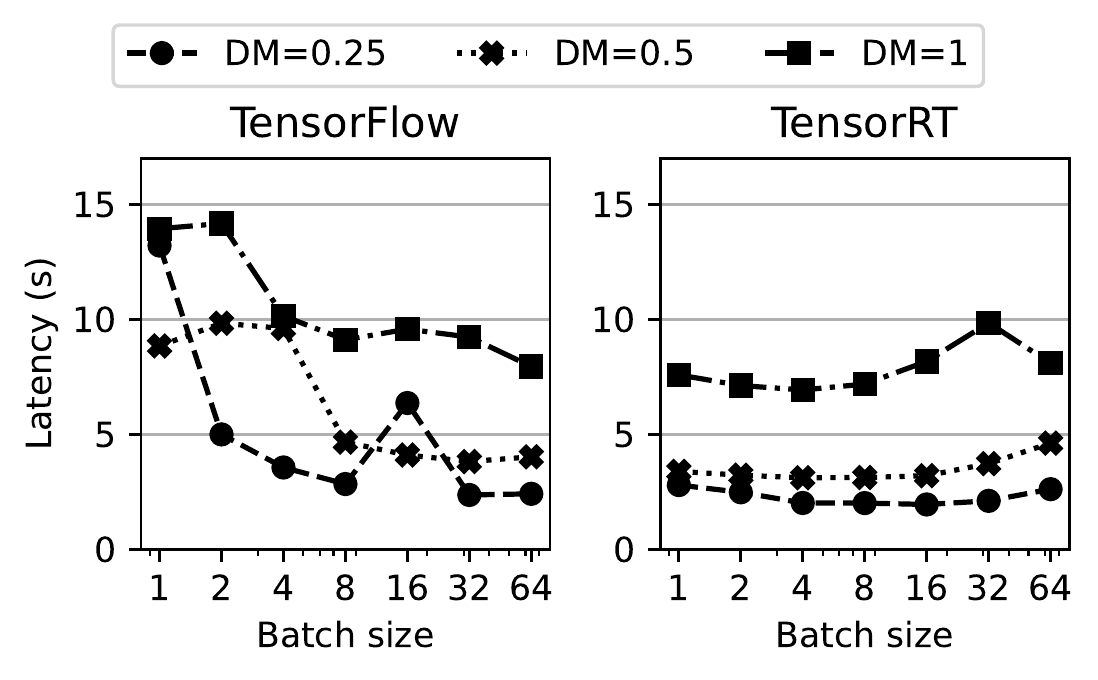}
\caption{Inference latency on NVIDIA Jetson Nano as the batch size vary while using TensorFlow and TensorRT. DM is the depth multiplier of MobileNetV1 to scale the model size.}
\label{fig:nano_latency}
\end{figure}

\subsubsection{NVIDIA Jetson Nano}\label{sec:ipu:optimizations:nano}

Modern GPUs can leverage multiple optimized alternative data types to the default 32-bit floating point data type. 
In the case of NVIDIA Jetson Nano, we take advantage of the 16-bit floating point data type,
which is the only alternative option on this device,
leading to an increase in performance and a decrease in memory requirements. 

\begin{figure}[]
\centering
\includegraphics[width=0.95\linewidth]{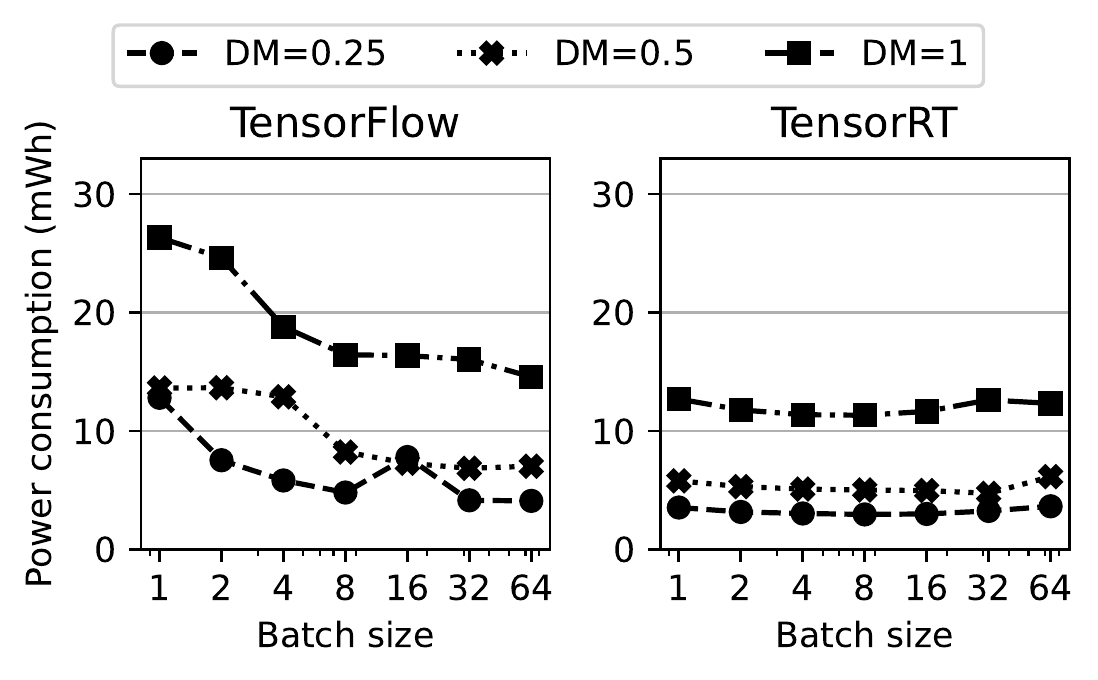}
\caption{Power consumption on NVIDIA Jetson Nano as the batch size vary while using TensorFlow and TensorRT. DM is the depth multiplier of MobileNetV1 to scale the model size.}
\label{fig:nano_power}
\end{figure}

This device is supported by TensorRT, a framework for building optimized inference engines for NVIDIA GPUs.
The optimizations this framework applies include layer and tensor fusion, kernel auto-tuning, dynamic tensor memory, etc.
This is also the only device in our setup that supports batching while performing inference. 
Therefore, we test the performance of the NVIDIA Jetson Nano using varying batch sizes and leveraging TensorFlow and TensorRT.
\Cref{fig:nano_latency} and \Cref{fig:nano_power} present the results.
They both clearly show the advantage of using the TensorRT framework.

The optimal batch size for inference using the TensorRT framework is 16, which achieves the best overall latency and power efficiency.
The increase in latency past this batch size is due to the total number of patches not being evenly divisible by these higher batch sizes, rendering a large portion of the computations of the last batch redundant. 
This problem only arises when using TensorRT, as the inference engines are compiled and optimized for a fixed batch size on this framework.

The division of the full-size image into patches and subsequent scaling and batching of these patches is performed by a separate thread using the \textit{numpy} library, pushing the processed batches into a queue.
The batches are then inferred using TensorFlow or TensorRT.
There are more optimized ways of creating the patches by reshaping the underlying numpy array of the whole image.
However, this leads to an increase in the latency of the first batch, affecting the overall latency, as this pre-processing step is performed concurrently with the inference.

\begin{figure}[h!]
\centering
\includegraphics[width=0.9\linewidth]{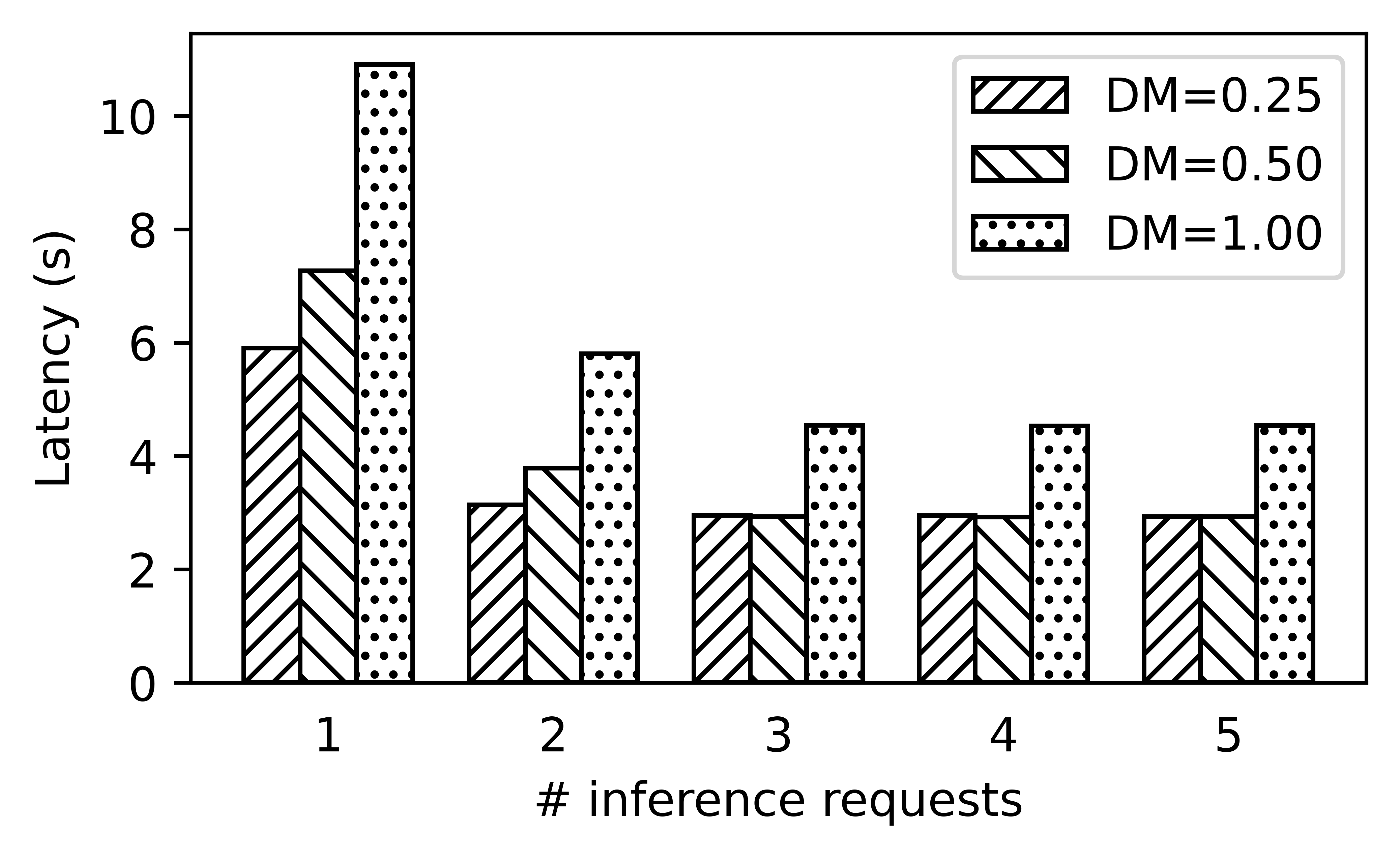}
\caption{Impact of the number of concurrent inference requests sent to the NCS2 on the inference latency. DM is the depth multiplier of MobileNetV1 to scale the model size.}
\label{fig:ncs_nireq}
\end{figure}

\subsubsection{Intel Neural Compute Stick 2}\label{sec:ipu:optimizations:ncs2}

While the NCS2 does not support batch sizes higher than one, it does support the submission of more than one inference request concurrently, with four requests being recommended by the manufacturer. 
These requests are then buffered on the device before their execution. 
This amortizes the time spent on data transfers between the NCS2 and the host, which is non-negligible, as shown in \Cref{fig:ncs_nireq}.

To saturate the device with data and therefore decrease the latency of inference for the entire image, we produce patches in a separate thread and push the processed images into a queue, as is the case for the NVIDIA Jetson Nano. 
While the NCS2 only supports the 16-bit floating point data type, the scaling of the input is integrated into the pre-processing steps of the model through OpenVINO.
To create the patches from the original image on the host CPU on the Raspberry Pi, we leverage the optimizations of the \textit{memcpy} operation to copy a full row of the patch matrix instead of copying every single value one at a time.

The main thread on Raspberry Pi polls the queue for new patches and populates the inference requests. 
As OpenVINO has support for asynchronous requests, the inference requests can be populated and submitted from the main thread to NCS2.
The inference requests accept a callback function, which is triggered on completion of inference, rendering the inference request available for reuse for inference request of the next patch.

\begin{figure}[h!]
\centering
\includegraphics[width=0.9\linewidth]{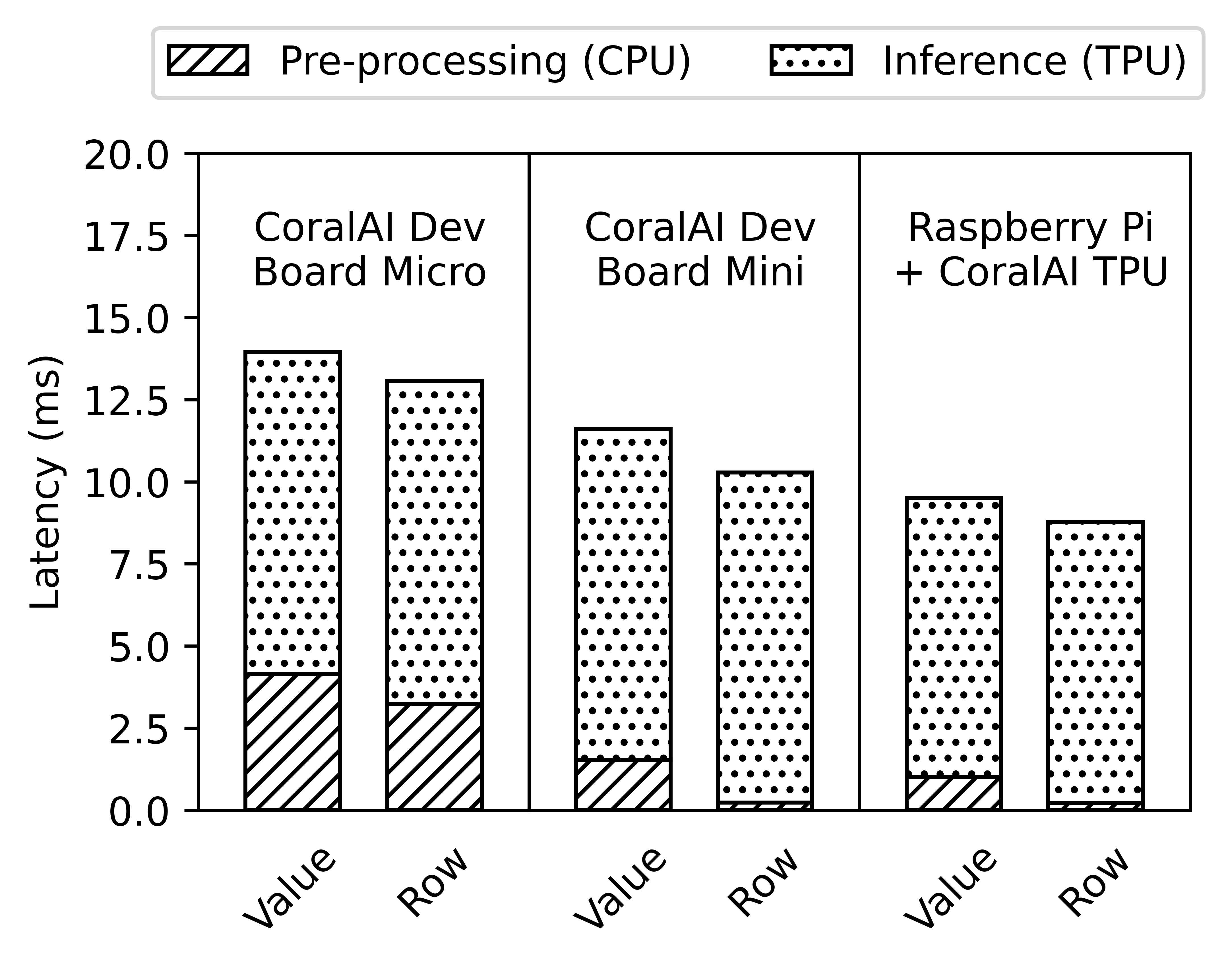}
\caption{Breakdown of the latency of the inference pipeline on the CoralAI Dev Board Micro, CoralAI Dev Board Mini, and the Raspberry Pi with the CoralAI USB accelerator. The latencies were measured on a single image patch using the model with a depth multiplier of 0.5. \textit{Value} bars show the version where the patches are created by copying one value at a time and the \textit{Row} bars show the optimized version that copies a whole row at a time using \textit{memcpy}.}
\label{fig:coral_breakdown}
\vspace{-1em}
\end{figure}

\subsubsection{CoralAI TPU}\label{sec:ipu:optimizations:tpu}

As mentioned in \Cref{sec:ipu:setup:devices}, the CoralAI TPU only supports the 8-bit integer data type. 
The network is, therefore, quantized to this precision. 
Unlike the NCS2, the TPU only supports synchronous requests. 

Since the precision of the network matches the data type of the image, there is no need for further pre-processing of the image after creating the patches. 
The creation of patches leverages the same \textit{memcpy} optimization as the NCS2. 
\Cref{fig:coral_breakdown} shows the breakdown of the latency for inference of a single image patch on all three setups leveraging the CoralAI TPU before and after applying this optimization.
The optimized version exhibits $1.28-6.71$x improvement in the latency of the pre-processing step (creation of the patch) of the inference pipeline.
While CoralAI Dev Board Micro exhibits the the lowest improvement by this optimization, the CoralAI Dev Board Mini exhibits the highest.

The creation of the patches on the CoralAI Dev Board Mini now accounts for only $91.2$ ms of the $4118$ ms of overall latency of inference of the whole image,
which is a small overhead.
Therefore, this pre-processing operation is run in the main thread as well, 
since the overhead of thread communication when we add a separate thread may give diminishing returns.

\begin{figure}[]
\centering
\includegraphics[width=\linewidth]{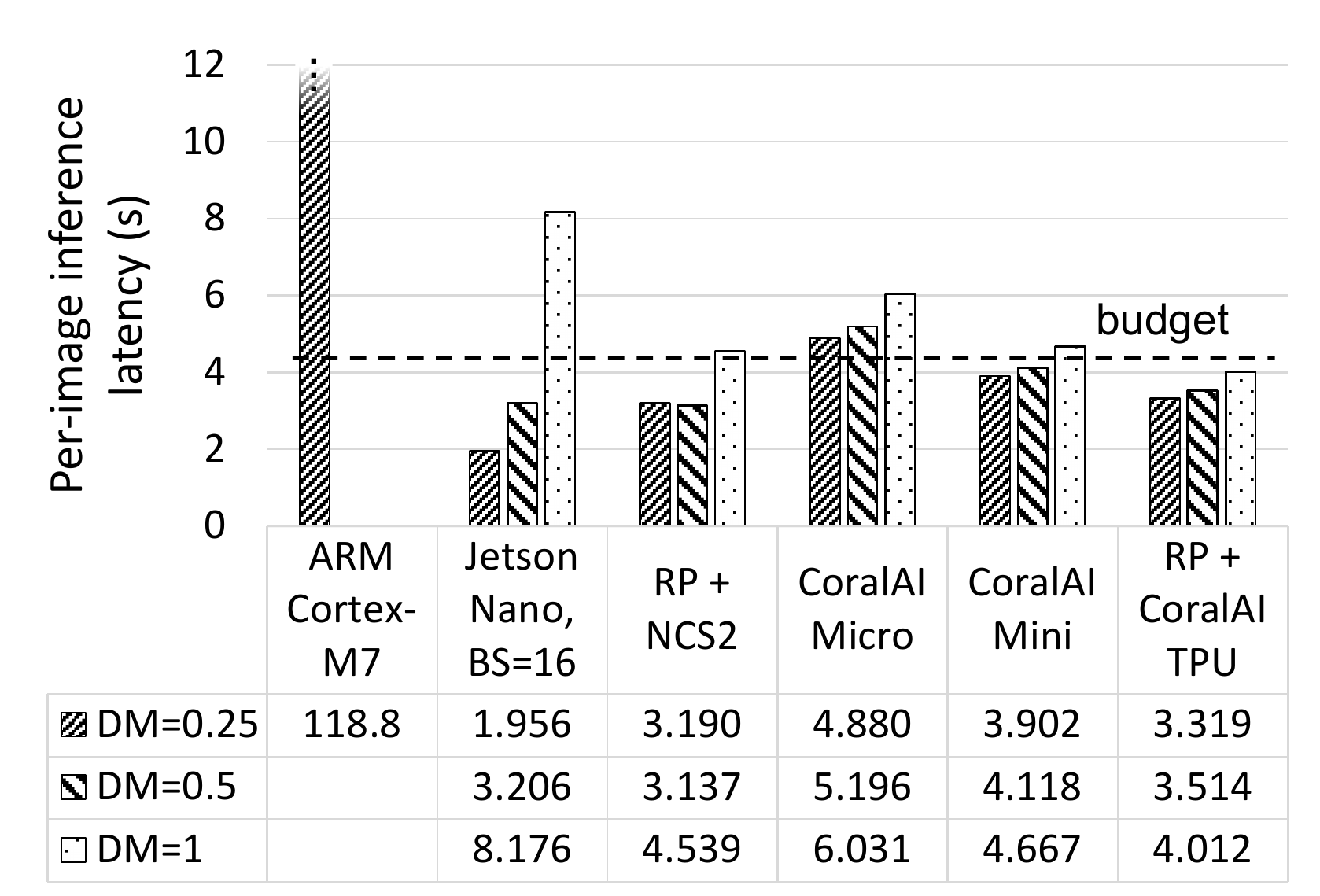}
\caption{Latency of inference on full-size image with different scaling factors (depth multiplier = DM) of the MobileNetV1 model as measured on corresponding devices. Batch size (BS) of 1 is used unless stated otherwise. The \textit{budget} line indicates the maximum latency budget for real-time imaging, i.e. \textit{scenario 1}, identified in \Cref{sec:requirements}.}
\label{fig:latency}
\end{figure}

\begin{figure}[]
\centering
\includegraphics[width=\linewidth]{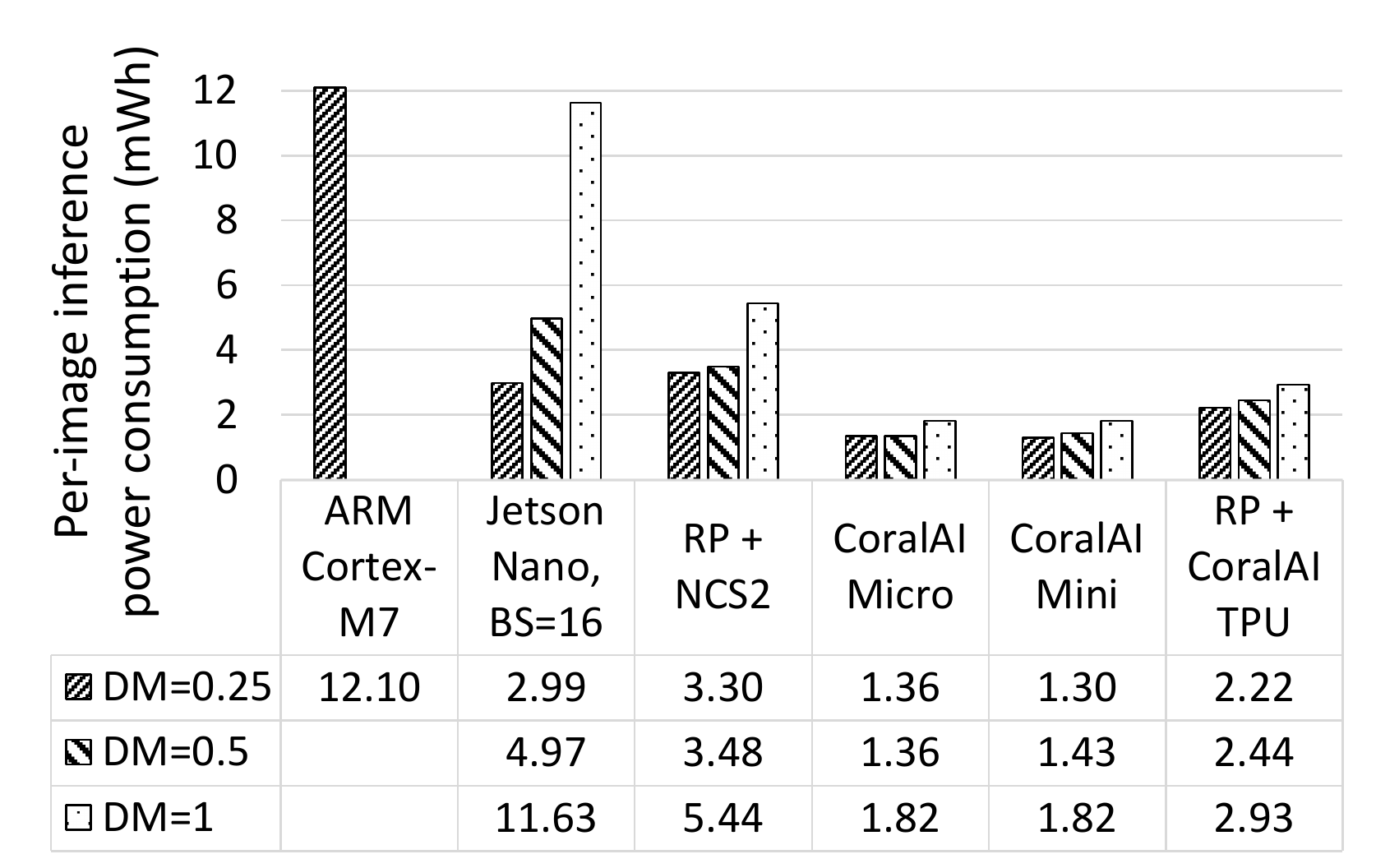}
\caption{Power consumption of inference on full-size image with different scaling factors (depth multiplier = DM) of the MobileNetV1 model as measured on corresponding devices. Batch size (BS) of 1 is used unless stated otherwise.}
\label{fig:power_consumption}
\end{figure}

\begin{figure*}[]
\centering
\includegraphics[width=\linewidth]{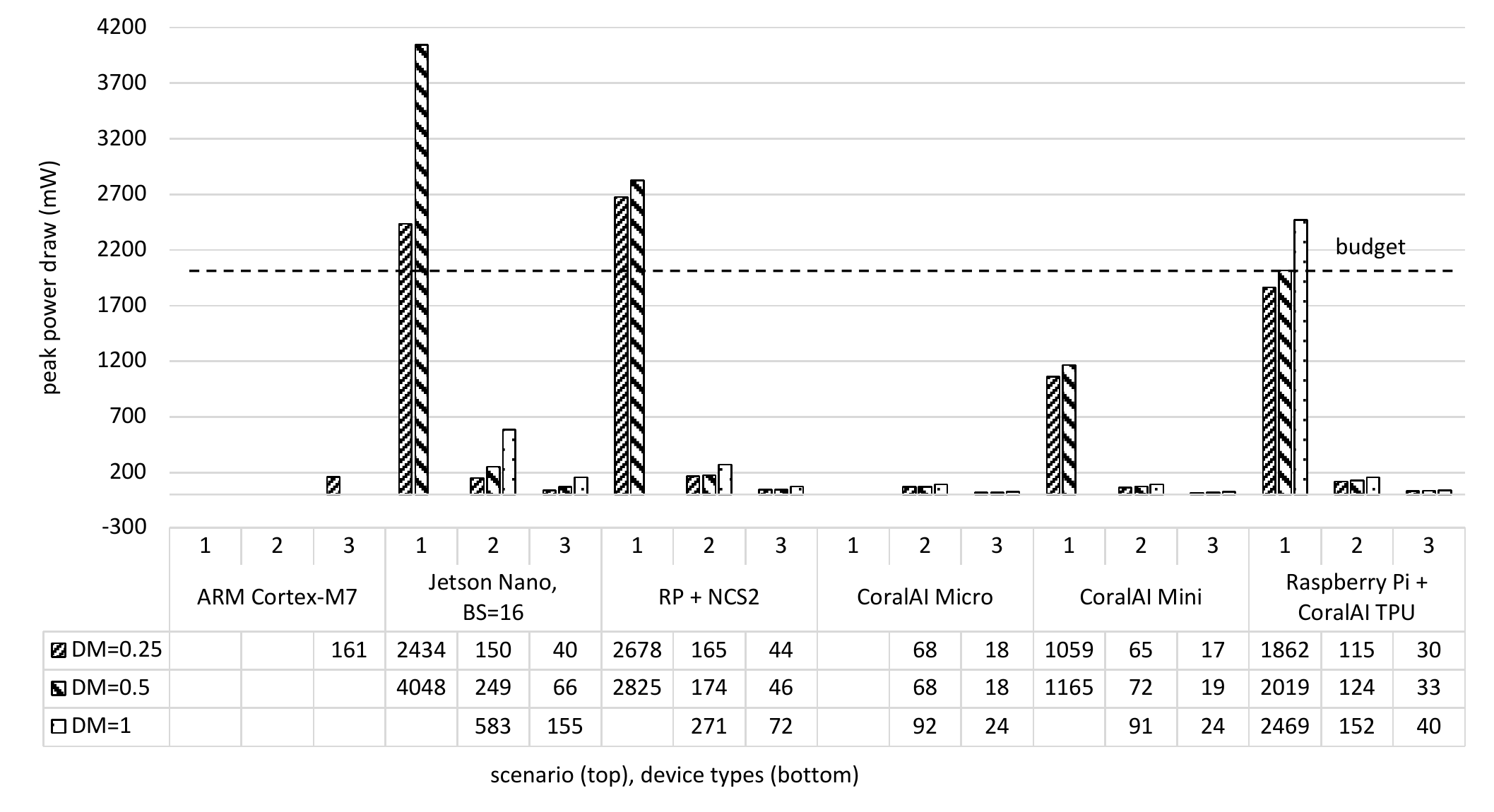}
\caption{Nominal power draw of each device with MobileNetV1 model of different sizes (depth multiplier = DM) for each scenario. The combinations of the devices and model sizes that do not fulfil the latency requirements are omitted as their active duty cycle is greater than 100\%. The \textit{budget} line marks the maximum nominal power allowed for the IPU as reported in \Cref{tab:constraints}.}
\label{fig:nominal_power}
\end{figure*}

\begin{figure}[h!]
\centering
\includegraphics[width=\linewidth]{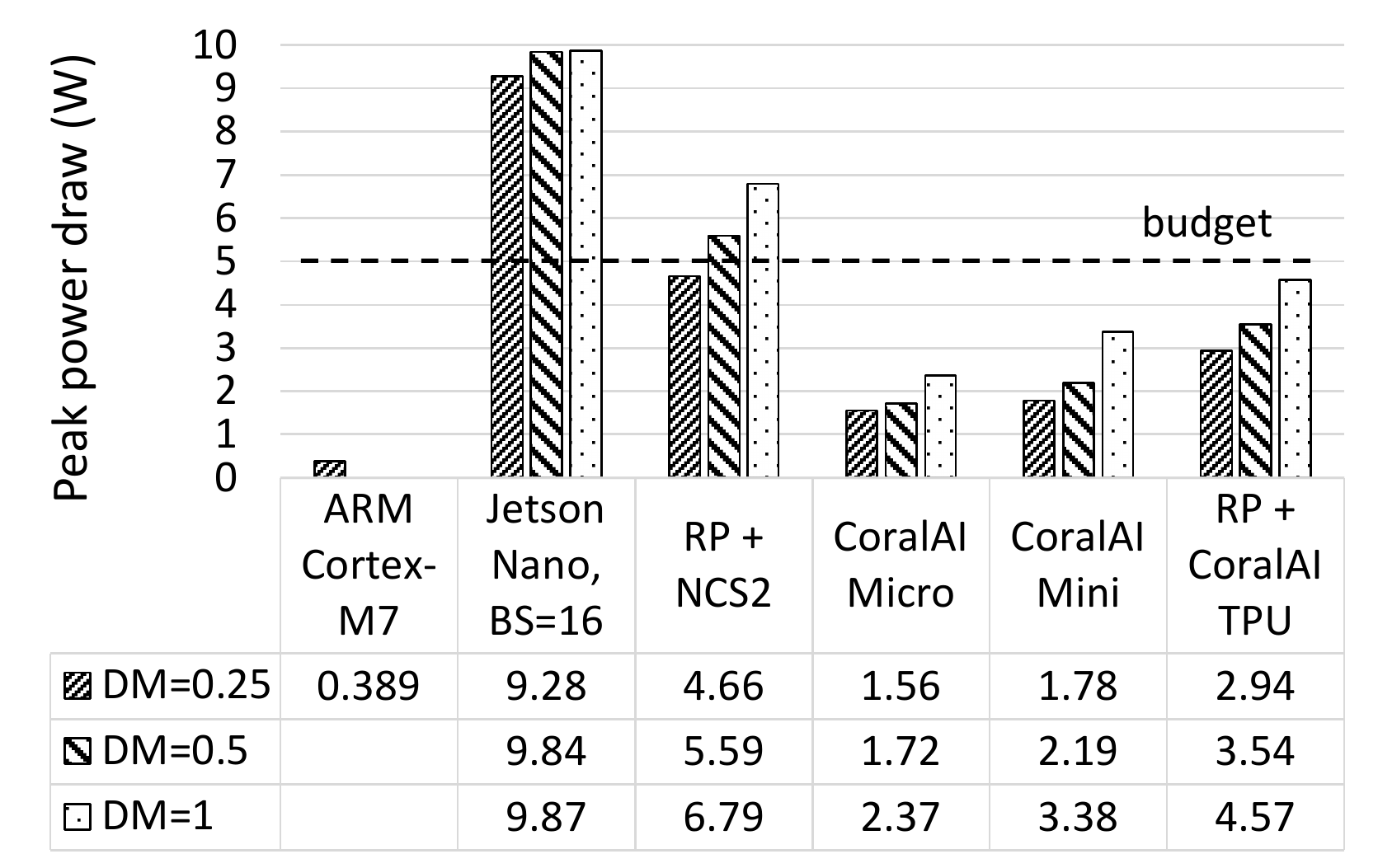}
\caption{Peak power draw of of each device with MobileNetV1 model of different sizes (depth multiplier = DM). Batch size (BS) of 1 is used if not stated otherwise. The \textit{budget} line indicates the maximum peak power allowed for the IPU as reported in \Cref{tab:constraints}.}
\label{fig:peak_power}
\end{figure}

\subsection{Comparison of Devices} \label{sec:ipu:results}

\Cref{fig:latency,fig:power_consumption} show the latency and the power consumption of inference using models with increasing size, i.e., with increased depth multiplier value.
Together with the nominal power draw in \Cref{fig:nominal_power} and peak power draw in \Cref{fig:peak_power}, 
\Cref{fig:latency,fig:power_consumption} guide our analysis of the suitability of these devices for use in imaging scenarios outlined in \Cref{sec:requirements}.
After the analysis with respect to each use case scenario, we show the effects of quantization and the depth multiplier on the models' accuracy and discuss implications for storage requirements.

\subsubsection{Scenario 1: Real-time imaging}
This is the most compute-intensive use case among the three scenarios. 
The low latency requirements also lead to a high duty cycle percentage causing the nominal power draw to approach the peak power draw of the devices that can fulfill the latency requirements of this scenario. 

\paragraph{Latency} 
As mentioned in \Cref{sec:requirements}, the pipeline has to achieve a $4.42$s latency for inference of the full-size image. 
The ARM Cortex-M7 MCU and the CoralAI Dev Board Micro are the only devices that cannot fulfill the latency requirements of this scenario
for any model size in \Cref{fig:latency}. 
This is caused by the lack of computing power of the CPU in these devices. 
Even though the CoralAI Dev Board Micro shares the same accelerator with the CoralAI Dev Board Mini and the Coral AI TPU USB accelerator, its CPU is unable to match the feed rate of the larger devices, underutilizing its TPU.
Overall, the NVIDIA Jetson Nano achieves latencies significantly lower than its counterparts in the two smaller models, while the latency of the largest model meets the latency requirements only in the case of Raspberry Pi with the CoralAI USB accelerator. 
We can see that compared to the rest of the devices, the setups utilizing the CoralAI TPU scale to larger models significantly better, best seen in the transition between the models with depth multipliers of 0.5 and 1.0.

\paragraph{Power draw} 
As mentioned earlier, the high active duty cycle leads to a high nominal power draw, causing multiple devices to exceed the power budget, highlighted in \Cref{fig:nominal_power}.
This means that out of the hardware devices that meet the latency requirements, only the CoralAI Dev Board Mini and the Raspberry Pi with the CoralAI USB accelerator fulfill the nominal power draw requirement at least for the small model configuration.
None of the devices, however, manage to fulfill all of the requirements for the largest model.

\subsubsection{Scenario 2: Arctic region imaging}
By relaxing the latency constraints via processing the images of only the areas above the arctic polar circle, we see more configurations passing the requirements necessary to perform such a workload.

\paragraph{Latency}
As \Cref{fig:latency} shows the relaxation of the compute requirements causes a significant portion of the devices to fulfill the latency requirements,
which is $71.74$s per image for this scenario (\Cref{sec:requirements}). 
The only device not fulfilling these requirements is the ARM Cortex-M7 MCU.
While the latency requirements are met for the rest of the hardware configurations, the power consumption or efficiency at which these devices can perform the inference varies (\Cref{fig:power_consumption}). 
The lowest power consumption is measured for both of the CoralAI Dev Boards, while the highest was measured on the NVIDIA Jetson Nano and the Raspberry Pi using the NCS2 accelerator, depending on the model size configuration.
The power consumption of the CoralAI dev boards is $26-84\%$ and $30-67\%$ lower than the power consumption of the NVIDIA Jetson Nano and the Raspberry Pi with the NCS2 accelerator, respectively.

\paragraph{Power draw}
As the majority of the setups, with the exception of the ARM MCU, either meet or are close to meeting the latency requirements of scenario 1, the active portion of the duty cycle in these devices decrease significantly, causing the nominal power draw to follow in 
\Cref{fig:nominal_power}.
All of these devices, therefore, fulfill the nominal power draw requirements in all model configurations.

\subsubsection{Scenario 3: Greenland imaging}
The least compute-demanding scenario corresponds to taking images of only Greenland. 
The low constraints mean that all of the devices pass the requirements under certain model configurations.

\paragraph{Latency}
This is the only scenario, which can be performed also using the low-power ARM Cortex-M7 MCU in addition to the other devices, since the latency requirement is $270$s per image. Even though this device has a very low power draw in comparison to the rest of the devices, the long latency to finish an inference makes it the least efficient choice among the devices we test, having a power consumption $3.67-9.31$x higher than its counterparts (\Cref{fig:power_consumption}). 

\paragraph{Power draw}
The majority of time in this scenario is spent idle and therefore the nominal power draw of the devices is minimal in \Cref{fig:nominal_power}.
With respect to the peak power draw shown in \Cref{fig:peak_power}, we see that all the configurations meet the requirement of less than 5W with two exceptions:
(1) NVIDIA Jetson Nano, whose power draw significantly exceeds the satellite's power budget in all model configurations, and
(2) the Raspberry Pi with the NCS2 accelerator, which fits within the power budget of the satellite only when using the smallest model.
Even though decreasing the batch size of the NVIDIA Jetson Nano decreases its peak power draw by up to 25.8\%, it does not lead to reductions significant enough to make this device eligible for deployment on the \DISCO~satellite.

\subsubsection{Effects of quantization and depth multiplier}
\Cref{tab:acc} shows the effect of post-training quantization on the validation accuracy of the models. 
Quantization of model parameters to 16-bit floating point values, shows only an insignificant loss of accuracy in the smallest model, while the accuracy of the two larger models remains exactly the same. 
Quantizing these models to 8-bit precision, disrupts the predictive performance of the model more significantly.
While the loss in accuracy is more than $2\%$ for the smallest model and
less than $0.5\%$ for the model with the depth multiplier of 0.5,
the largest model exhibits an opposite effect, where the validation accuracy increases from $90.74\%$ up to $91.55\%$ with 8-bit precision.

Furthermore, \Cref{tab:acc} shows the number of parameters of each of the models.
We deem the model with the depth multiplier of 0.5 the most efficient.
On the one hand, it shows significant improvement in accuracy in comparison to the smallest model.
On the other hand, it does not suffer a big accuracy drop compared to the largest model while exhibiting a significant decrease in the number of parameters.

\begin{table}
    \centering
    \begin{tabular}{@{}llrrr@{}}
    \toprule
       & & \multicolumn{3}{c}{Depth multiplier} \\
    & Precision   & \multicolumn{1}{c}{0.25} & \multicolumn{1}{c}{0.5} & \multicolumn{1}{c}{1.0}        \\
    \midrule
    \multirow{3}{*}{Accuracy}
    & 32 bit float  
        & 86.92\%  
        & 90.33\%  
        & 90.74\% \\ 
    & 16 bit float
        & 86.78\% 
        & 90.33\% 
        & 90.74\% \\ 
    & 8 bit integer 
        & 84.33\% 
        & 89.78\% 
        & 91.55\% \\ 
    \midrule
    \multicolumn{2}{c}{\# params}
    & 219,829 
    & 832,101 
    & 3,233,989 \\
    \bottomrule
    \end{tabular}
    \caption{Accuracy of the MobilenetV1 model with different model sizes as the precision varies (full-precision vs quantizatied) as measured on the Flowers dataset, and the number of parameters of the different model sizes.}
    \vspace{-3em}
    \label{tab:acc}
\end{table}

\subsubsection{Storage}
Even though the benchmarking results do not consider storage capacity and the I/O speed, they are very important, mainly for the scenarios 2 and 3. 
In order to increase the required latency compared to the real-time scenario of 4.42 seconds, the streaming frequency of the camera needs to be decreased to stream the images to storage, to be later read and inferred.
In order to buffer the 80 or 320 images outlined for the scenarios 2 and 3, respectively, in \Cref{sec:requirements}, the devices have to have $80*60MB=4,800MB$ and $320*60MB=19,200MB$ of storage, respectively. 
This is fulfilled by all the device setups, with the exception of CoralAI Dev Board Mini (fulfills only the lower requirement) and Micro.
On the other hand, both of these devices have support for an SD card expansion,
which can easily solve the storage space issue.

\subsection{Summary of Results} \label{sec:ipu:summary}

In \Cref{sec:ipu:results},
the CoralAI-TPU-based devices showed the most promising results as they were the only ones to fulfill the latency and power requirements for real-time imaging, scenario 1, which is the most challenging scenario. 
The device is highly specialized for neural network inference, utilizing the systolic array architecture. 
This architecture is purpose-built to perform fast multiply-accumulate operations in a highly parallel fashion allowing for performing matrix multiplication without the need to load/store intermediate values, leading to superior efficiency.

At the other end of the spectrum was the ARM Cortex-M7 MCU.
Due to the lack of parallelism, the device could not fulfill the latency requirements of any use case scenarios, except for the least challenging one.
The low peak power draw did not lead to increased efficiency in comparison to the other devices, due to the high latency. 
Furthermore, this device and the CoralAI Dev Board Micro have far less memory, not allowing the whole image to fit in memory. 
This can be overcome with in-camera cropping and heavy use of buffering to storage, which would decrease efficiency and increase the complexity of the solution even further.
To report latency results on the ARM Cortex-M7 MCU and the CoralAI Dev Board Micro, we use images of total sizes of 224 x 224 (single patch) and 2272 x 2272 (100 patches) pixels, respectively. The results of the benchmarks for these subsets are extrapolated to 400 patches by multiplying the latency by 400 and 4, respectively, in the case of the ARM Cortex-M7 MCU and CoralAI Dev Board Micro.

While the NVIDIA Jetson Nano and the Raspberry Pi paired with the NCS2 can match or even improve on the latency of the CoralAI-TPU-based devices, they possess worse power characteristics, exceeding the budget for peak and nominal power draw.

\section{The Onboard IPU Module}\label{sec:system}

Guided by the results presented in \Cref{sec:ipu:results},
this section first illustrates the hardware architecture of the IPU module of the \DISCO~satellite in \Cref{sec:system:hardware},
after which it proceeds by describing the software running on the module in \Cref{sec:system:software}.

\subsection{Hardware}\label{sec:system:hardware}
\Cref{sec:ipu:results} demonstrates that the highest degree of specialization is needed to perform the highly compute-demanding workload of classification of satellite imagery while fulfilling the resource constraints such as the highly limited power budget. 
While multiple devices could fit into the latency requirements of the most resource-demanding use case scenario using the model with the depth multiplier of 0.5, the CoralAI Dev Board Mini offers the most power-efficient setup, being the only device that has a satisfying nominal power draw of less than 2W.

The CoralAI Dev Board Mini makes the IPU the most computationally powerful module on the satellite.
This is not only due to the TPU accelerator but also its comparatively large CPU with four cores and 2GB of LPDDR3 RAM.
This host CPU runs Mendel OS \cite{mendelos}, Debian-based Linux distribution, and exposes a variety of I/O, including I2C, SPI, UART, MIPI CSI, and multiple GPIO pins.

\begin{figure}[h]
\centering
\includegraphics[width=0.95\linewidth]{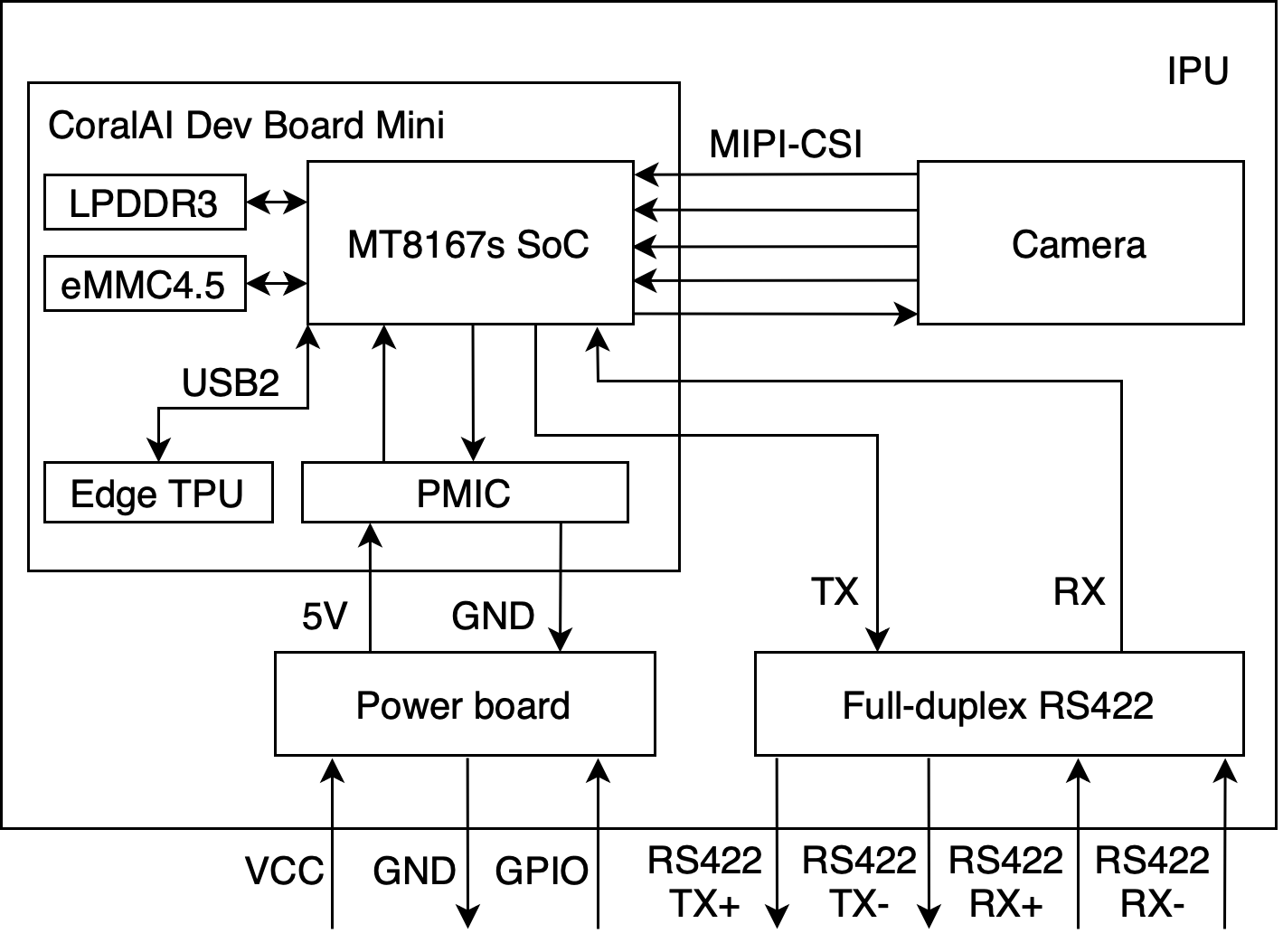}
\caption{Block diagram of the hardware architecture of the Image Processing Unit (IPU) module on \DISCO.}
\label{fig:ipu}
\vspace{-1.5em}
\end{figure}

As \Cref{fig:ipu} illustrates, the IPU module is composed of multiple other components contained inside an aluminium enclosure.
We will now introduce each of them.

\paragraph{Power board}
The satellite's PCDU module
has multiple power delivery channels, which can operate independently of each other. 
However, the mission-critical modules are given priority when dividing th limited number of channels between the modules (\Cref{sec:background:disco:modules}).
This meant that it was necessary for the IPU module to share a channel with the OBC module.
This affected the supply voltage, which was initially set to match the voltage of the OBC module at 9V.
This value was too high for the CoralAI Dev Board Mini, and therefore the supply voltage had to be stepped down to 5V. 
To enable this, the board leverages a high-efficiency switching voltage regulator.
Furthermore, the channel sharing meant that the two modules had to operate simultaneously. 
This was, however, unsustainable, as the CoralAI Dev Board Mini and the peripherals draw a significant amount of power compared to other modules (up to 2W of average power draw, compared to 0.9W of average power draw of the PCDU and the OBC combined), even when idle. 
Therefore, this board also includes a circuit for switching the IPU on/off using GPIO signals from the OBC, enabling the OBC to run without the overhead of running the IPU module when needed.

\paragraph{Full-duplex RS422}
As mentioned in \Cref{sec:background:disco:arch}, CSP accepts multiple network interfaces, with the most common being the CAN bus and RS422 using the KISS protocol. 
Our module takes advantage of RS422 because of the readily available UART interface on the CoralAI Dev Board Mini, which RS422 depends on. 
The introduction of the CAN interface would require the inclusion of an SPI-based CAN controller accompanied by the required kernel module and the CAN transceiver.
As the other modules do not expose the RS422 interface, by using a point-to-point connection between the OBC and IPU modules we introduce a single point of failure, which could be eliminated through the use of the CAN bus.
However, as we described in the previous paragraph, the OBC is responsible for switching this module on/off, making the  OBC and its connection to the IPU a single point of failure in any case.

\paragraph{Camera}
A small camera is included inside the aluminium enclosure, in order to conduct cosmic ray\footnote{High-energy particles, which can cause damage to electronics upon collision with the satellite.} detection experiments. 
By taking long-exposure images in total darkness, there is a high probability that a high-energy particle hits the camera's CCD array and leaves a spot or a streak of high-intensity pixels in the image, depending on the angle of the incidence. 
These experiments will enable us to quantify the number of occurrences of cosmic ray collisions and correlate these with possible outages of the system. 

\subsection{Software}\label{sec:system:software}

As mentioned in \Cref{sec:system:hardware}, the CoralAI Dev Board Mini runs a custom Linux distribution based on Debian called Mendel OS. 
On the one hand, this made the development of the software for this device convenient through the use of amenities such as SSH sessions and a file system. 
On the other hand, the inclusion of an operating system comes at the cost of increased complexity, which makes the identification of possible failures very hard. 
Operating systems provide a high level of abstraction and perform many background tasks, including writes to the persistent storage. 
This leads to a high possibility of system corruption after the loss of power.

This fragile behavior is unacceptable for deployment in remote areas, especially satellites, which have to be highly resilient.
To cope with this, we mount the partitions of the file system as read-only, except for the home partition, which hosts all of the custom software and stores the images, logs of experiments, and the IPU module's parameter table.

In addition to the OS, the software on the IPU includes the CSP and the services we expose on this module, which are:

\paragraph{FTP server}
The FTP server was developed primarily to send images from the satellite down to the ground station. 
This service, nevertheless, works on any file type and in the other direction as well. 
In addition to downloading, the service exposes and facilitates uploading, listing, moving, copying, and removing files. 
This enables us to additionally upload files, including scripts and models to update the data analysis and processing tasks on the satellite, and download files other than images, such as various log files for post-inspection of events.
The upload and download of large files have a high probability of being interrupted, with some taking multiple days. 
The FTP server divides these larger files into smaller chunks, which are sent individually and error checked before being joined back. 
These files can for example be the compiled MobileNet models, which are 384KB, 1.1MB, and 3.5MB larger for depth multipliers of 0.25, 0.5, and 1, respectively
on the CoralAI device.

\paragraph{Inference test}
We deploy a benchmark running the MobileNetV1 model with a depth multiplier of 0.5 on a dataset with golden truth logits. 
This will let us not only track the performance of the CoralAI Dev Board Mini, specifically its onboard TPU, but also detect possible faults in the real-world satellite setup. 
This benchmark also logs the latencies and the logits of the inference together with the timestamps
to help making correlations between data collection and other events such as the cosmic ray detections.

\paragraph{Cosmic ray detection}
This service uses the small camera inside the enclosure to take images in order to perform cosmic ray detection. 
This uses the Otsu's thresholding algorithm \cite{otsu} to detect the high-intensity clusters of pixels.
Once detected, these can then be cropped and sent back to a ground station, thereby also testing the data pipeline for sending images.
Alternatively, one can populate the log files with these detections in order to make a correlation with possible faults in the systems caused by the impact of these high-energy particles.

\paragraph{Telemetry}
In addition to collecting metrics related to the workloads, such as the inference test or the cosmic ray detection, the IPU also exposes the live telemetry data, such as the CPU utilization, memory usage, disk space available, or the temperature of the CPU, through the CSP's parameter tables. 
These can be streamed from the satellite together with other sources of telemetry such as the power draw of the different modules from the PCDU. 

\paragraph{User-defined workload}
Despite the Arctic satellite imagery being the main use case of the \DISCO~satellite, the IPU module also has to support other student-driven image analysis and processing projects. 
These projects are to be launched while the satellite is in space, and therefore we need a place to define the workload as a service and couple it with the CSP to expose them to the user.
While the rest of the services are rigidly coupled with CSP and compiled into a single program, this is an infeasible approach to upload these student projects onto the IPU module due to the limited bandwidth between the satellite and the ground station. 
We, therefore, expose a service that can run arbitrary user-defined workloads, without the need to update the entire set of software services.

While every new user-defined workload will go through comprehensive testing on the on-ground replica of the satellite, we implement a slot system for ensuring further robustness guarantees. 
We define four slots for the software updates on satellite with the satellite booting into the next slot on every startup or after a watchdog timer elapses.
This way when encountering an error after uploading a new software into a slot, the satellite can revert to a known state, even if it becomes unreachable due to the upload.


\subsection{Discussion and Future Work} \label{sec:system:discussion}

%
%

\paragraph{Toolchains.}
Compared to other solutions, the devices leveraging the CoralAI TPU to accelerate the deep learning workloads use a more closed-source toolchain. 
The CoralAI compiler is closed-source, meaning the optimizations to the inference using this device can only stem from high-level optimizations within either the scope of the TensorFlow Lite framework or the code running on the host CPU.
While the solution offers great out-of-the-box performance, it does not leave much room for additional contributions from the research community.

FPGAs could bridge the gap between the performance and availability of open-source solutions by leveraging frameworks such as FINN \cite{finn} or hls4ml \cite{hls4ml}.
The choice of an FPGA-based IPU was however discarded after a qualitative trade-off analysis at the beginning of the project.
The reasons were the short time frame for development of the IPU, the large upload sizes connected with updates of the bitstream on FPGAs, 
and the concerns related to failure modes of FPGA reconfigurations after deployment.

\paragraph{Network.}
While CSP promotes a distributed architecture and no-single-point-of-failure approach, sharing of the power channel between the OBC and the IPU modules prevented us from truly achieving these features on \DISCO1. 
For \DISCO2 we aim to have these modules on separate channels and connect the IPU module on the CAN BUS, 
in addition to including a faster Ethernet connection to the radio module to increase the speed at which the IPU can offload images from the satellite. 
The inclusion of Ethernet on the satellite, would also ease the operation of cameras and enable multiple modules to operate the cameras.
This way the IPU could be responsible only for processing the images, with the possibility of passing the raw images to other modules in case of IPU failure.

\paragraph{Model updates.}
The images are inferred on the satellite to enable the smart utilization of the limited bandwidth between the satellite and the ground station, by only sending images relevant to the mission at hand. 
This approach, however, prevents us from building a balanced dataset for future training. While the CoralAI TPU has the ability to retrain the last layer of the neural network, the retraining can only be performed using the images sent to a ground station.
These images can be relabeled on the ground station, and these new labels then can be sent back to the satellite. 
If the initial model was trained correctly, this will lead to a dataset heavily skewed towards the images deemed important for the mission.
Therefore, we also need to dedicate some of the bandwidth to send at least highly compressed versions of the images deemed unimportant. 
The low resolution will prevent us from using the images to build a new dataset, but the newly acquired labels could be used for fine-tuning the model on board of the satellite.

Even though the MobileNetV1 models are considered very small in size by today's standards, their size is large enough to put high pressure on the link between the satellite and a ground station. 
There is a need for research to compress these networks in transit or applications of partial updates.
The MobileNetV1 can, for example, be dissected to detach the feature extractor from the head of the network (the last fully-connected layer), which would allow us to update this last layer to fit a new use case at hand while saving a significant portion of the bandwidth. 
These two parts of the model can then be reassembled and compiled for use for multiple use cases on the TPU on board of the satellite.

\section{Conclusion}
\label{sec:conc}

This paper presented our work and lessons-learned on building an Image Processing Unit (IPU) for a satellite in the \DISCO~project.
We first characterized the performance of six systems that were possible candidates for accelerating deep-learning-based image processing on board of a small satellite.
Our results demonstrate the necessity of using highly specialized architecture.
The latency requirements combined with the limited power, mass, and size budget of our use cases were fulfilled by only one device, the CoralAI Dev Board Mini.
While, the NVIDIA Jetson Nano and the Raspberry Pi with the NCS2 accelerator matched or even improved upon the latency of the CoralAI Dev Board Mini, their low power-efficiency in comparison to the CoralAI-TPU-based devices, leads to exceeding the power budget of our satellite.
We then detailed how we used the Coral AI Dev Board Mini in combination with other components to create an IPU module and integrated this module into the distributed architecture of the \DISCO~satellites.
The end-product was launched in space in April 2023 and will soon serve as a testbed for several student projects and future satellite designs for similar Earth observation use cases.


\bibliographystyle{ACM-Reference-Format}
\bibliography{sample-base}

\end{document}